\documentclass{article}
\usepackage{microtype}
\usepackage{graphicx}
\usepackage{subfigure}
\usepackage{booktabs} 
\usepackage{hyperref}
\usepackage{amsmath,amssymb}
\usepackage{authblk}
\usepackage[accepted]{icml2020}
\newcommand{\hideh}[1]{}
\usepackage{amsmath,amssymb,amsthm,xspace}

\newtheorem{theorem}{Theorem}[section]
\newtheorem{definition}[theorem]{Definition}
\newtheorem{assumption}[theorem]{Assumption}
\newtheorem{lemma}[theorem]{Lemma}
\newtheorem{proposition}[theorem]{Proposition}

\newcommand{\X}{\mathcal{X}}
\newcommand{\Y}{\mathcal{Y}}

\newcommand{\Finterp}{\mathcal{E}}
\newcommand{\ROPE}{RObust Post hoc Explanations\xspace}
\newcommand{\rope}{ROPE\xspace}
\newcommand{\E}{\mathbb{E}}

\newcommand{\argmax}{\operatorname*{\arg\max}}
\newcommand{\argmin}{\operatorname*{\arg\min}}
\newcommand{\hide}[1]{}

\icmltitlerunning{Robust and Stable Black Box Explanations}

\begin{document}

\twocolumn[
\icmltitle{Robust and Stable Black Box Explanations}
\begin{icmlauthorlist}
\icmlauthor{Himabindu Lakkaraju}{har}
\icmlauthor{Nino Arsov}{mac}
\icmlauthor{Osbert Bastani}{penn}
\end{icmlauthorlist}

\icmlaffiliation{har}{Harvard University}
\icmlaffiliation{mac}{Macedonian Academy of Arts \& Sciences}
\icmlaffiliation{penn}{University of Pennsylvania}

\icmlcorrespondingauthor{Himabindu Lakkaraju}{hlakkaraju@hbs.edu}

\vskip 0.3in
]

\printAffiliationsAndNotice{}
\begin{abstract}
As machine learning black boxes are increasingly being deployed in real-world applications, there has been a growing interest in developing post hoc explanations that summarize the behaviors of these black boxes. However, existing algorithms for generating such explanations have been shown to lack stability and robustness to distribution shifts. We propose a novel framework for generating robust and stable explanations of black box models based on adversarial training. Our framework optimizes a minimax objective that aims to construct the highest fidelity explanation with respect to the worst-case over a set of adversarial perturbations. We instantiate this algorithm for explanations in the form of linear models and decision sets by devising the required optimization procedures. To the best of our knowledge, this work makes the first attempt at generating post hoc explanations that are robust to a general class of adversarial perturbations that are of practical interest. Experimental evaluation with real-world and synthetic datasets demonstrates that our approach substantially improves robustness of explanations without sacrificing their fidelity on the original data distribution.
\end{abstract}

\section{Introduction}\label{sec:intro}
Over the past decade, 
there has been an increasing interest in leveraging machine learning (ML) models to aid decision making in critical domains such as healthcare and criminal justice. However, the successful adoption of these models in the real world relies heavily on how well decision makers are able to understand and trust their functionality \cite{doshi2017towards,lipton2016mythos}.
Decision makers must have a clear understanding of the model behavior so they can diagnose errors and potential biases in these models, and decide when and how to employ them. However, the proprietary nature and increasing complexity of machine learning models poses a severe challenge to understanding these complex \emph{black boxes}, motivating the need for tools that can explain them in a faithful and interpretable manner.

Several different kinds of approaches have been proposed to produce interpretable \emph{post hoc explanations} of black box models. For instance, LIME and SHAP~\cite{ribeiro16:kdd,lundberg2017unified} explain individual predictions of any given black box classifier via \emph{local} approximations. On the other hand, approaches such as MUSE~\cite{lakkaraju19faithful} focus on explaining the high-level \emph{global} behavior of any given black box.

However, recent work has shown that post hoc explanation methods are \emph{unstable} (i.e., small perturbations to the input can substantially change the constructed explanations), as well as \emph{not robust to distribution shifts} (i.e., explanations constructed using a given data distribution may not be valid on others)~\cite{ghorbani2019interpretation,lakkaraju2020how}. 
A key reason why many post hoc explanation methods are not robust is that they construct explanations by optimizing fidelity on a given covariate distribution $p(x)$~\cite{ribeiro2018anchors,ribeiro16:kdd,lakkaraju19faithful}---i.e., choose the explanation that makes the same predictions as the black box on $p(x)$. 
To see why these approaches may fail to be robust,
consider a covariate distribution $p(x_1,x_2)$ where $x_1$ and $x_2$ are perfectly correlated, and an outcome $y=\mathbb{I}[x_1\ge0]$. Suppose we have a black box $B^*(x_1,x_2)=\mathbb{I}[x_2\ge0]$, and an explanation $\hat{E}(x_1,x_2)=\mathbb{I}[x_1\ge0]$. Since $x_1$ and $x_2$ are perfectly correlated, the explanation has perfect fidelity---i.e.,
\begin{align}
\label{eqn:ex}
\mathbb{P}_{p(x_1,x_2)}[\hat{E}(x_1,x_2)=B^*(x_1,x_2)]=1.
\end{align}
Thus, $\hat{E}$ appears to be a good explanation of $B^*$. However, if the underlying covariate distribution changes---e.g., to $p'(x_1,x_2)$ where $x_1$ and $x_2$ are independent---then $\hat{E}$ no longer has high fidelity. 

The lack of robustness is problematic because many of the undesirable behaviors of black box models that can be diagnosed using interpretability relate to distribution shifts. For instance, it has been shown that interpretability can help users in assessing whether a model would transfer well to a new domain~\cite{ribeiro16:kdd}---e.g., from one hospital to another~\cite{bastani2018predicting}; ~\citet{caruana15:intelligible} show that experts use interpretable models to identify spurious relationships which do not hold if the underlying data changes---e.g., if a patient has asthma, he is not likely to die from pneumonia; these are intrinsically distribution shift issues. 
Thus, for the explanations to shed light on these kinds of issues in the black box, high fidelity on the original distribution alone may be insufficient; instead, it also needs to achieve high fidelity on the relevant shifted distributions. To further complicate the problem, we often do not know in advance what are the relevant distribution shifts. Therefore, constructing explanations that are robust to a general class of possible shifts is of great importance.

We propose a novel algorithmic framework, \ROPE (\rope) for constructing black box explanations that are not only stable but also robust to shifts in the underlying data distribution. To the best of our knowledge, our work is the first attempt at generating robust post hoc explanations for black boxes. \rope focuses on two notions of robustness. The first is \emph{adversarial robustness}~\cite{ghorbani2019interpretation}, which intuitively says that if the inputs are adversarially perturbed (by small amounts), then the explanation should not change significantly. The second is \emph{distributional robustness}~\cite{namkoong2016stochastic}, which is similar to adversarial robustness but considers perturbations to the input distribution rather than individual inputs.
While \rope considers distributional and adversarial robustness, these properties also improve stability. This is due to the fact that explanations designed to be robust to input perturbations are not likely to vary drastically with small changes in inputs. 

First, we propose a novel minimax objective that can be used to construct robust black box explanations for a given family of interpretable models. This objective encodes the goal of returning the highest fidelity explanation with respect to the worst-case over a set of distribution shifts.

Second, we propose a set of distribution shifts that captures our intuition about the kinds of shifts to which interpretations should be robust. In particular, this set includes shifts that contain perturbations to a small number of covariates. For instance, robustness to these shifts ensure that the marginal dependence of the black box on a single covariate is preserved in the explanation, since the explanation must be robust to changes in that covariate alone.

Third, we propose algorithms for optimizing this objective in two settings: (i) explanations such as linear models with continuous parameters that can be optimized using gradient descent, in which case we use adversarial training~\cite{goodfellow15:explaining}, and (ii) explanations such as decision sets with discrete parameters, in which case we use a sampling-based approximation in conjunction with submodular optimization~\cite{lakkaraju16:interpretable}. 

We evaluated our approach \rope on real-world data from healthcare, criminal justice, and education, focusing on datasets that include some kind of distribution shift---i.e., individuals from two different subgroups (e.g., patients from two different counties). Our results demonstrate that the explanations constructed using \rope are substantially more robust to distribution shifts than those generated by state-of-the-art post hoc explanation techniques such as LIME, SHAP, and MUSE. Furthermore, the fidelity of \rope explanations is equal or higher than the fidelity of the explanations generated by state-of-the-art methods even on the original data distribution, thus demonstrating that \rope improves robustness of explanations without sacrificing their fidelity on the original data distributions. In addition, we used synthetic data to analyze how the degree of distribution shift affects fidelity of the explanations constructed by our approach and other baselines. Finally, we performed an experiment where the ``black box'' models are themselves interpretable, and showed that \rope explanations constructed based on shifted data are substantially more similar to the black box than the explanations output by other baselines.

\section{Related Work}
\textbf{Post hoc explanations.}
Many approaches have been proposed to directly learn interpretable models~\cite{breiman2017classification,tibshirani1997lasso,letham15:interpretable,lakkaraju16:interpretable,caruana15:intelligible,kim2019learning}; however, complex models such as deep neural networks and random forests typically achieve higher accuracy than simpler interpretable models~\cite{ribeiro16:kdd}; thus, it is often desirable to use complex models and then construct post hoc explanations to understand their behavior.

A variety of post hoc explanation techniques have been proposed, which differ in their access to the complex model (i.e., black box vs. access to internals), scope of approximation (e.g., global vs. local), search technique (e.g., perturbation-based vs. gradient-based), explanation families (e.g., linear vs. non-linear), etc. For instance, in addition to LIME~\cite{ribeiro16:kdd} and SHAP~\cite{lundberg17:a-unified}, 
several other \emph{local explanation} methods have been proposed that compute \emph{saliency maps} which capture importance of each feature for an individual prediction by computing the gradient with respect to the input~\cite{simonyan2013saliency, sundararajan2017axiomatic, selvaraju2017grad,smilkov2017smoothgrad}. 
An alternate approach is to provide a global explanation summarizing the black box as a whole~\cite{lakkaraju19:faithful,bastani2017interpretability}, typically using an interpretable model. 

There has also been recent work on exploring vulnerabilities of black box explanations~\cite{adebayo2018sanity,slack2019can,lakkaraju2020how,rudin2019stop,dombrowski2019explanations}---e.g., \citet{ghorbani2019interpretation} demonstrated that post hoc explanations can be unstable, changing drastically even with small perturbations to inputs. However, none of the prior work has studied the problem of constructing robust explanations.

\textbf{Distribution shift.}
\emph{Distribution shift} refers to settings where there is a mismatch between the training and test distributions. A lot of work in this space has focused on \emph{covariate shift}, where the covariate distribution $p(x)$ changes but the outcome distribution $p(y\mid x)$ remains the same. This problem has been studied in the context of learning predictive models~\cite{Candela2009,jiang2007}. Proposed solutions include importance weighting~\cite{shimodaira2000improving}, invariant representation learning~\cite{ben2007analysis,tzeng2017adversarial}, online learning~\cite{cesa2006prediction}, and learning adversarially robust models~\cite{teo2007convex,herbrich2004invariant,decoste2002training}. However, none of these approaches are applicable in our setting since they assume either that the underlying predictive model is not a black box, that data from the shifted distribution is available, or that the black box can be adaptively retrained.

\textbf{Adversarial robustness.}
Due to the discovery that deep neural networks are not robust~\cite{szegedy2014intriguing}, there has been recent interest in \emph{adversarial training}~\cite{goodfellow15:explaining,bastani2016measuring,sinha2018certifying,shaham2018understanding}, which optimizes a minimax objective that captures the worst-case over a given set of perturbations to the input data. At a high level, these algorithms are based on gradient descent; at each gradient step, they solve an optimization problem to find the worst-case perturbation, and then compute the gradient at this perturbation. For instance, for $L_{\infty}$ robustness (i.e., perturbations of bounded $L_{\infty}$ norm), ~\citet{goodfellow15:explaining} propose to approximate the optimization problem using a single gradient step, called the \emph{signed-gradient} update; ~\citet{shaham2018understanding} generalizes this approach to arbitrary norms. We propose a set of perturbations that capture our intuition about the kinds of distribution shifts that explanations should be robust to; for this set of shifts, we show how approximations along the lines of these previous approaches correspond to solving a linear program on every step to compute the gradient.

\section{Our Framework}
Here, we describe our framework for constructing robust explanations. We assume we are given a black box model $B^*:\X\to\Y$, where $\X\subseteq\mathbb{R}^n$ is the space of covariates and $\Y$ is the space of labels. Our goal is to construct a global explanation for the computation performed by $B^*$. To construct such an explanation, one approach would be to learn an interpretable model that approximates $B^*$. In particular, given a family $\Finterp$ of interpretable models, a distribution $p(x)$ over $\X$, and a loss function $\ell:\Y\times\Y\to\mathbb{R}$, this approach constructs an explanation as follows:
\begin{align}
\label{eqn:explanation}
\vspace{-0.4in}
\hat{E}(x)=\argmin_{E\in\Finterp}\mathbb{E}_{p(x)}[\ell(E(x),B^*(x))].
\vspace{-0.4in}
\end{align}
In other words, $\hat{E}$ minimizes the error (as defined by $\ell$)
relative to the black box $B^*$. Intuitively, if $\hat{E}$ is a good approximation of $B^*$, then the computation performed by $B^*$ should be mirrored by the computation performed by $\hat{E}$.

The problem with Eq.~\ref{eqn:explanation} is that it only guarantees that $\hat{E}$ is a good approximation of $B^*$ according to the distribution $p(x)$. If the underlying data distribution changes, then $\hat{E}$ may no longer be a good approximation of $B^*$.

\subsection{Robust \& Stable Explanations}

To construct explanations that are robust to shifts in the data distribution $p$, we first consider the general setting where we are given a set of distribution shifts that we want our explanations to be robust to; we describe a practical choice in Section~\ref{sec:deltachoice}. We initially focus on distributional robustness; we connect it to adversarial robustness below.
\begin{definition}
\rm
Let $p$ be a distribution over $\X$, and let $\delta\in\mathbb{R}^n$. The \emph{$\delta$-shifted distribution} is $p_{\delta}(x)=p(x-\delta)$.
\end{definition}
In other words, $p_{\delta}$ places probability mass on covariates that are shifted by $\delta$ compared to $p$.
\begin{definition}
\rm
Let $p$ be a distribution over $\X$. Given $\Delta\subseteq\mathbb{R}^n$, the set of \emph{$\Delta$-small shifts} is the set $\{p_{\delta}\mid\delta\in\Delta\}$ of $\delta$-shifted distributions.
\end{definition}
For computational tractability, we assume:
\begin{assumption}
\rm
\label{assump:delta}
The set $\Delta$ of shifts is a convex polytope.
\end{assumption}
Given a set of distribution shifts, our goal is to compute the best explanation that is robust to these shifts:
\begin{definition}
\rm
Given $\Delta\subseteq\mathbb{R}^n$, $s_0\in\mathbb{N}$, and $\delta_{\text{max}}\in\mathbb{R}_{>0}$, the \emph{optimal robust explanation} for $(s_0,\delta_{\text{max}})$-small shifts is
\begin{align}
\label{eqn:distrobust}
\hat{E}&=\argmin_{E\in\Finterp}\max_{\delta\in\Delta}\mathbb{E}_{p_{\delta}(x)}[\ell(E(x),B^*(x))].
\end{align}
\end{definition}
That is, $\hat{E}$ optimizes the worst-case loss over shifts $p_{\delta}$. Computing the worst-case over shifts $p_{\delta}$ can be intractable; instead, we use an upper bound on the objective in Eq.~\ref{eqn:distrobust}.
\begin{lemma}
\label{lem:surrogate}
We have
\begin{align*}
&\max_{\delta\in\Delta}\mathbb{E}_{p_{\delta}(x)}[\ell(E(x),B^*(x))] \\
&\le\mathbb{E}_{p(x)}\left[\max_{\delta\in\Delta}\ell(E(x+\delta),B^*(x+\delta))\right].
\end{align*}
\end{lemma} 
\emph{Proof:} Note that
\begin{align*}
&\max_{\delta\in\Delta}\mathbb{E}_{p_{\delta}(x)}[\ell(E(x),B^*(x))] \\
&=\max_{\delta\in\Delta}\int_{\X}\ell(E(x),B^*(x))p(x-\delta)dx \\
&=\max_{\delta\in\Delta}\int_{\X}\ell(E(x'+\delta),B^*(x'+\delta))p(x')dx' \\
&\le\int_{\X}\max_{\delta\in\Delta}\ell(E(x'+\delta),B^*(x'+\delta))p(x')dx',\\
&\le\mathbb{E}_{p(x)}\left[\max_{\delta\in\Delta}\ell(E(x+\delta),B^*(x+\delta))\right] \qed \end{align*} 

This lemma gives us a surrogate objective that we can optimize in place of the one in Eq.~\ref{eqn:distrobust}---i.e.,
\begin{align}
\label{eqn:advrobust}
\hat{E}=\argmin_{E\in\Finterp}\mathbb{E}_{p(x)}\left[\max_{\delta\in\Delta}\ell(E(x+\delta),B^*(x+\delta))\right].
\end{align}
In particular, this approach connects distributional robustness to adversarial robustness---Eq.~\ref{eqn:advrobust} is the standard objective used to achieve adversarial robustness to input perturbations $\delta\in\Delta$~\cite{goodfellow15:explaining}.

\subsection{General Class of Distribution Shifts}
\label{sec:deltachoice}

Next, we propose a choice of $\Delta$ that captures distributions shifts we believe to be of importance in practical applications. We begin with a concrete setting that motivates our choice, but our choice includes shifts beyond this setting.

In particular, consider the case where $\X=\{0,1\}^d$ is a vector of indicators. Our intuition is that when examining an explanation, users often want to understand how the model predictions change when a handful of components of an input $x\in\X$ change. 

For instance, this intuition captures the case of counterfactual explanations, where the goal is to identify a small number of covariates that can be changed to affect the outcome~\cite{zhang2018interpreting}. It also captures certain intuitions underlying fairness and causality, where we care about how the model changes when a covariate such as gender or ethnicity changes~\cite{lakkaraju2020how,rosenbaum1983central,pearl2009causality}. 
Finally, it also encompasses the shifts considered in measures of variable importance~\cite{hastie01:the-elements}---in particular, variable importance measures how the explanation changes when a \emph{single} component of the input $x$ is changed.

We can use the following choice to capture our intuition:
\begin{align*}
\Delta_1=\{\delta\in\{-1,0,1\}^n\mid\|\delta\|_0\le s_0\}
\end{align*}
for $s_0\in\mathbb{N}$. However, this set is nonconvex. We can approximate this constraint using the following set:
\begin{align*}
\Delta_2=\{\delta\in\mathbb{R}^n\mid\|\delta\|_0\le s_0\wedge\|\delta\|_{\infty}\le1\}.
\end{align*}
In particular, the constraint $\|\delta\|_{\infty}\le1$ ensures that $-1\le \delta_i\le 1$ for each $i\in\{1,...,n\}$. Finally, we can replace the $L_0$ norm with the $L_1$ norm:
\begin{align}
\label{eqn:exdelta}
\Delta_3=\{\delta\in\mathbb{R}^n\mid\|\delta\|_1\le s_0\wedge\|\delta\|_{\infty}\le1\}.
\end{align}
This overapproximation is a heuristic based on the fact that the $L_1$ loss induces sparsity in regression~\cite{tibshirani1997lasso}.

More generally, we consider a shift from $p$ to a distribution $p'$ such that $p'$ places probability mass on the same inputs $x$ as $p$, except a small number of components of $x$ are systematically changed by a small amount:
\begin{align*}
\tilde{\Delta}(s_0,\delta_{\text{max}})=\{\delta\in\mathbb{R}^n\mid\|\delta\|_0\le s_0\wedge\|\delta\|_{\infty}\le\delta_{\text{max}}\},
\end{align*}
where $s_0\in\mathbb{N}$ and $\delta_{\text{max}}\in\mathbb{R}$---i.e., $\delta\in\tilde{\Delta}(s_0,\delta_{\text{max}})$ is a sparse vector whose components are not too large. However, $\tilde{\Delta}(s_0,\delta_{\text{max}})$ is nonconvex. As above, for computational tractability, we approximate it using
\begin{align*}
\Delta(s_0,\delta_{\text{max}})=\{\delta\in\mathbb{R}^n\mid\|\delta\|_1\le s_0\wedge\|\delta\|_{\infty}\le\delta_{\text{max}}\}.
\end{align*}
It is easy to see that $\tilde{\Delta}(s_0,\delta_{\text{max}})\subseteq\Delta(s_0,\delta_{\text{max}})$, so this choice overapproximates the set of shifts. In particular, this choice $\Delta(s_0,\delta_{\text{max}})$ is a polytope, so it satisfies Assumption~\ref{assump:delta}. The set defined in Eq.~\ref{eqn:exdelta} is $\Delta_3=\Delta(s_0,1)$.

A particular benefit of $\Delta(s_0,\delta_{\text{max}})$ is that the marginal dependencies of $B^*$ on a component $x_i$ of an input $x\in\X$ is preserved in $\hat{E}$---i.e., if we unilaterally change $x_i$ by a small amount, $B^*$ and $\hat{E}$ change in the same way. Formally:
\begin{proposition}
\label{prop:delta}
Suppose $\Y=\mathbb{R}$, $\ell(y,y')=|y-y'|$, and $\Delta=\Delta(s_0,\delta_{\text{max}})$, and consider an explanation $\hat{E}$ with error
\begin{align*}
\mathbb{E}_{p(x)}\left[\max_{\delta\in\Delta}\ell(\hat{E}(x+\delta),B^*(x+\delta))\right]\le\epsilon.
\end{align*}
Then, letting $\alpha$ be the the one-hot encoding of $i$ (i.e., $\alpha_i=1$ and $\alpha_j=0$ if $i\neq j$), for any $c\in\mathbb{R}$ such that $|c|\le\delta_{\text{max}}$,
\begin{align*}
&\mathbb{E}_{p(x)}\left[\left|(\hat{E}(x+c\alpha)-\hat{E}(x))-(B^*(x+c\alpha)-B^*(x))\right|\right] \\
&\hspace{2.8in}\le2\epsilon.
\end{align*}
\end{proposition}
\vspace{-0.20in}
\emph{Proof}: Note that
\begin{align*}
&\mathbb{E}_{p(x)}\left[\left|(\hat{E}(x+c\alpha)-\hat{E}(x))-(B^*(x+c\alpha)-B^*(x))\right|\right] \\
&\le\mathbb{E}_{p(x)}\left[\left|\hat{E}(x+c\alpha)-B^*(x+c\alpha)\right|\right] \\
&\hspace{0.2in}+\mathbb{E}_{p(x)}\left[\left|\hat{E}(x)-B^*(x)\right|\right] \\
&\le2\epsilon \text{ since } c\alpha\in\Delta \qed
\end{align*}
As shown in Section~\ref{sec:intro}, this property is not satisfied by standard measures of fidelity, since an explanation with perfect fidelity (i.e., Eq.~\ref{eqn:ex}) may use completely different covariates from the black box.

\subsection{Constructing Robust Linear Explanations}

We consider the case where $\Finterp$ is the space of linear functions, or more generally, any model family that can be optimized using gradient descent. Then, we can use \emph{adversarial training} to optimize Eq.~\ref{eqn:advrobust}~\cite{goodfellow15:explaining,shaham2018understanding}. The key idea behind adversarial training is to learn a model $f^* \in \mathcal{F}$ that is robust 
with respect to a worst-case set of perturbations to the input data---i.e.,
\begin{align*}
f^*=\argmin_{f\in\mathcal{F}}\mathbb{E}_{p(x,y)}\left[\max_{\delta\in\Delta}\ell(f(x+\delta),y)\right].
\end{align*}
We can straightforwardly adapt this formalism to our setting by replacing $\mathcal{F}$ with $\Finterp$ and $y$ with $B^*(x)$.
In particular, suppose that $E_\theta\in\Finterp$ is parameterized by $\theta\in\Theta$, where $\Theta\subseteq\mathbb{R}^d$ and $J(\theta;x)$ is defined as follows:
\begin{align*}
J(\theta;x)=\ell(E_\theta(x),B^*(x)).
\end{align*}
Then, Eq.~\ref{eqn:advrobust} becomes
\begin{align}
\label{eqn:advrobustJ}
\hat{\theta}=\argmin_{\theta\in\Theta}\E_{p(x)}\left[\max_{\delta\in\Delta}J(\theta;x+\delta))\right].
\end{align}
The adversarial training approach optimizes Eq.~\ref{eqn:advrobustJ} by using stochastic gradient descent~\cite{goodfellow15:explaining,shaham2018understanding}---for a single sample $x\sim p(x)$, the stochastic gradient estimate of the objective in Eq.~\ref{eqn:advrobustJ} is
\begin{align*}
\nabla_{\theta}\max_{\delta\in\Delta}J(\theta;x+\delta)\approx\nabla_\theta J(\theta;x+\delta^*),
\end{align*}
where
\vspace{-0.20in}
\begin{align}
\label{eqn:delta}
\delta^*=\argmax_{\delta\in\Delta}J(\theta,x+\delta).
\end{align}
To solve Eq.~\ref{eqn:delta}, we use the Taylor approximation
\begin{align*}
J(\theta;x+\delta)\approx J(\theta;x)+\nabla_xJ(\theta;x)^\top\delta.
\end{align*}
Using this approximation, Eq.~\ref{eqn:delta} becomes
\begin{align}
\delta^*&=\argmax_{\delta\in\Delta}J(\theta,x+\delta) \nonumber \\
&\approx\argmax_{\delta\in\Delta}\left\{J(\theta;x)+\nabla_xJ(\theta;x)^\top\delta\right\} \nonumber \\
&=\argmax_{\delta\in\Delta}\nabla_xJ(\theta;x)^\top\delta, \label{eqn:deltataylor}
\end{align}
where in the last line, we dropped the term $J(\theta;x)$ since it is constant with respect to $\delta$. Since we have assumed $\Delta$ is a polytope, Eq.~\ref{eqn:deltataylor} is a linear program with free variables $\delta$. 

\subsection{Constructing Robust Rule-Based Explanations}

Here, we describe how we can construct robust rule-based explanations~\cite{lakkaraju16:interpretable,letham15:interpretable,lakkaraju19faithful}---e.g., decision sets~\cite{lakkaraju16:interpretable,lakkaraju19faithful}, decision lists~\cite{letham15:interpretable}, decision trees~\cite{quinlan1986induction}. Any rule based model can be expressed as a decision set~\cite{lakkaraju2017learning}, so we focus on these models.

Unlike explanations with continuous parameters, we can no longer use gradient descent to optimize Eq.~\ref{eqn:advrobust}. Instead, we optimize it using a sampling-based heuristic. We assume we are given a distribution $p_0(\delta)$ over shifts $\delta\in\Delta$. Then, we approximate the maximum in Eq.~\ref{eqn:advrobust} using $k$ samples:
\begin{align*}
\max_{\delta\in\Delta}F(\delta)\approx\max_{\delta^j\sim p_0(\delta)}F(\delta^j),
\end{align*}
where $F(\delta)$ is a general objective and $j\in\{1,...,k\}$. In particular, our optimization problem becomes
\begin{align}
\label{eqn:samplerobust}
\hat{E}=\argmin_{E\in\Finterp}\mathbb{E}_{p(x)}\left[\max_{\delta^j\sim p_0(\delta)}\ell(E(x+\delta^j),B^*(x+\delta^j))\right].
\vspace{-0.3in}
\end{align}
\vspace{-0.1in}
Next, a decision set
\begin{align*}
E = \{(s_1,c_1), (s_2,c_2) \cdots (s_m, c_m)\}\subseteq S\times C
\end{align*}
is a set of rules of the form $(s,c)$ where $s$ is a conjunction of predicates of the form $(\text{feature}, \text{operator}, \text{value})$ (e.g., age $\geq$ 45) and $c\in\Y$ is a label.
Typically, we consider the case where $\Y$ is a finite set. Existing algorithms~\cite{lakkaraju19faithful,lakkaraju16:interpretable} for constructing decision set explanations primarily optimize for the following three goals: (i) maximizing the \emph{coverage} of $E$---i.e., for $x\in\X$, maximizing the probability that one of the rules $(s,c)\in E$ has a condition $s$ that is satisfied by $x$, (ii) minimizing the \emph{disagreement} between $E$ and $B^*$---i.e., minimizing the probability that $E(x)\neq B^*(x)$, and (iii) minimizing the complexity of $E$---e.g., $E$ has fewer rules. In particular, these algorithms optimize the following objective:
\begin{align}
\label{eqn:opti}
\hat{E}=&\argmax_{E \subseteq S \times C}
\{-\text{disagree}(E) + \lambda \cdot \text{cover}(E)\} \\
&\text{subj. to } |E| \leq \alpha, \nonumber \end{align}
where
\vspace{-0.15in}
\begin{align*}
\text{disagree}(E) &= \sum_{i=1}^m \mathbb{P}_{p(x)}(s_i(x)\rightarrow B^*(x) \neq c_i) \\
\text{cover}(E) &= \mathbb{P}_{p(x)}(\exists(s,c)\in E\text{ s.t. }s(x)=\text{true}).
\end{align*}
Here, we let $s(x)=\text{true}$ if $x$ satisfies $s$ and $s(x)=\text{false}$ otherwise. In $\text{disagree}(E)$, the event in the probability says if predicate $s_i$ applies to $x$, then $B^*(x) \neq c_i$. 

To adapt this approach to solving Eq.~\ref{eqn:samplerobust}, we modify the disagreement to take the worst-case over $\delta^j\sim p_0(\delta)$:
\begin{align*}
&\text{disagree}(E) \\
&=
\sum_{i=1}^m\mathbb{P}_{p(x)}\left(s_i(x)\rightarrow\exists\delta^j\sim p_0(\delta)~.~B^*(x+\delta^j)\neq c_i\right).
\end{align*}
where $j\in\{1,...,k\}$. Here, we have used an approximation where we only check if $s_i$ applies to the unperturbed input $x$; this choice enables our submodularity guarantee.
\begin{theorem}
\label{thm:submodular}
Suppose that $p(x)=\text{Uniform}(X_{\text{train}})$, where $X_{\text{train}}\subseteq\X$ is a training set, is the empirical training distribution. Then, the optimization problem Eq.~\ref{eqn:opti} is non-monotone and submodular with cardinality constraints.
\end{theorem}
\vspace{-0.18in}
\emph{Proof:} To show non-monotonicity, it suffices to show that at least one term in the objective Eq.~\ref{eqn:opti} is non-monotone. Every time a new rule is added, the value of disagree either remains the same or increases, since the newly added rule may potentially label new instances incorrectly, but does not decrease the number of instances already labeled incorrectly by previously chosen rules. Therefore, $\text{disagree}(A) \leq \text{disagree}(B)$ if $A \subseteq B$, so $-\text{disagree}(A) \geq-\text{disagree}(B)$, which implies disagree term is non-monotone. Thus, the entire linear combination is non-monotone.
\\
To prove that the objective in Eq.~\ref{eqn:opti} is submodular, we need to: (i) introduce a (large enough) constant $C$ into the objective function to ensure that $C - \text{disagree}(E)$ is never negative,\footnote{Note that adding such a constant does not impact the solution to the optimization problem.} and (ii) prove that each of the its terms are submodular. The \emph{cover} term is clearly submodular---i.e., more data points will be covered when we add a new rule to a smaller set of rules compared to a larger set. It is also easy to check that the \emph{disagree} term is modular/additive (and therefore submodular). 
Lastly, the constraint in Eq.~\ref{eqn:opti} is a cardinality constraint. $\qed$ \\
\\
Since the objective of Eqn.~\ref{eqn:opti} is non-monotone and submodular with cardinality constraints (Theorem 3.7), exactly solving it is NP-Hard~\cite{khuller1999budgeted}. So, we use approximate local search algorithm~\cite{lee2009non} to optimize Eq.~\ref{eqn:opti}. This algorithm provides the best known theoretical guarantees for this class of problems---i.e., $(k+2+1/k+\delta)^{-1}$, where $k$ is the number of constraints ($k=1$ in our case) and $\delta > 0$.

\section{Experiments}
\begin{table*}
\small
\centering
\begin{tabular}{l c c c c c c c c c}
    \toprule
        \bf Algorithms & \multicolumn{3}{c}{\bf Bail} & \multicolumn{3}{c}{\bf Academic} & \multicolumn{3}{c}{\bf Health} \\ \midrule
    & Train & Shift & \% Drop & Train & Shift & \% Drop & Train & Shift & \% Drop\\
    \midrule
    LIME & 0.79 & 0.64 & 18.99\% & 0.68 & 0.57 & 16.18\% & 0.81 & 0.69 & 14.81\% \\
    SHAP & 0.76 & 0.66 & 13.16\% & 0.67 & 0.59 & 11.94\% & 0.83 & 0.68 & 18.07\%\\
    MUSE & 0.75 & 0.59 & 21.33\% & 0.66 & 0.51 & 22.73\% & 0.79 & 0.61 & 22.78\%\\
    \midrule
    \rope logistic & 0.61 & 0.59 & {\bf 3.28\%} & 0.57 & 0.57 & {\bf 0.00\%} & 0.70 & 0.68 & {\bf 2.86\%} \\
    \rope dset & 0.64 & 0.61 & 4.69\% & 0.65 & 0.63 & 3.08\% & 0.73 & 0.69 & 5.48\%\\
    \midrule
    \rope logistic multi & 0.79 & 0.74 & 6.33\% & 0.70 & 0.69 & 1.43\% & 0.82 & 0.76 & 7.32\%\\
    \rope dset multi & {\bf 0.82} & {\bf 0.77} & 6.1\% & {\bf 0.73} & {\bf 0.71} & 2.74\% & {\bf 0.84} & {\bf 0.78} & 7.14\% \\
    \bottomrule
    \end{tabular}
\caption{Fidelity values of all the explanations are reported on both training data and shifted data, along with percentage drop in fidelity from training data to shifted data. Smaller values of percentage drop correspond to more robust explanations.}
\vspace{-0.1in}
\label{tab:distshift}
\end{table*}

As part of our evaluation, we first use
real-world data to assess the robustness of the post hoc explanations constructed using our algorithm and compare it to state-of-the-art baselines. Second, on synthetic data, we analyze how varying the degree of distribution shift impacts the fidelity of our explanations. Third, we ascertain the correctness of explanations generated using our framework---in particular, in cases where the black box is also an interpretable model $B^*\in\Finterp$, we study how closely the constructed explanations resemble the ground truth black box model. 

\subsection{Experimental Setup}

\textbf{Datasets.}
We analyzed three real-world datasets from criminal justice, healthcare, and education domains~\cite{lakkaraju16:interpretable}. Our first dataset contains \textbf{bail outcomes} from two different state courts in the U.S. 1990-2009. It includes criminal history, demographic attributes, information about current offenses, and other details on 31K defendants who were released on bail. Each defendant in the dataset is labeled as either high risk or low risk depending on whether they committed new crimes when released on bail. Our second dataset contains \textbf{academic performance} records of about 19K students who were set to graduate high school in 2012 from two different school districts in the U.S. It includes information about grades, absence rates, suspensions, and tardiness scores from grades 6 to 8 for each of these students. Each student is assigned a class label indicating whether the student graduated high school on time. Our third dataset contains \textbf{electronic health records} of about 22K patients who visited hospitals in two different counties in California between 2010-2012. It includes demographic information, symptoms, current and past medical conditions, and family history of each patient. Each patient is assigned a class label which indicates whether the patient has been diagnosed with diabetes.

\textbf{Distribution shifts.}
Each of our datasets contains two different subgroups---e.g., our bail outcomes dataset contains defendants from two different states. We randomly choose data from one of these subgroups (e.g., a particular state) to be the \emph{training data}, and data from the other subgroup to be the \emph{shifted data}. In particular, we apply each algorithm on the training data to construct explanations, and evaluate these explanations on the shifted data.

\textbf{Our explanations.}
Our framework \rope can be applied in a variety of configurations. We consider four: (i) \rope logistic: We construct a single global logistic regression model using our framework to approximate any given black box. (ii) \rope dset: We construct a single global decision set using our framework to approximate any given black box. (iii) \rope logistic multi: We construct multiple \emph{local} explanations. In particular, we first cluster the data into $K$ subgroups (details below), and use \rope to fit a robust logistic regression model to approximate the given black box for each subgroup. We also compute the centroid of each subgroup to serve as a representative sample. (iv) \rope dset multi: Similar to \rope logistic multi, except that we fit a decision set.

\begin{figure*}
\centering
\includegraphics[width=0.31\textwidth]{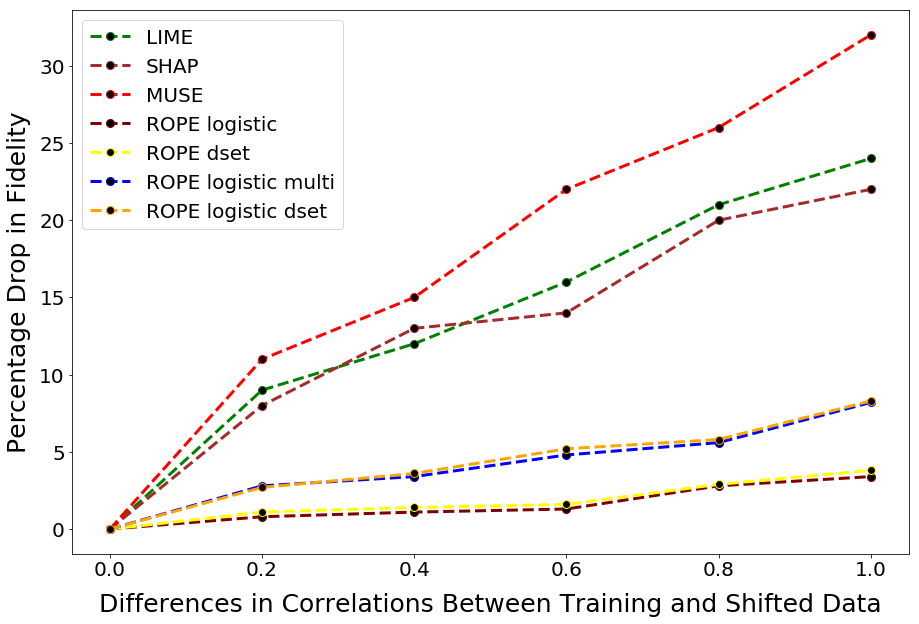}
\includegraphics[width=0.31\textwidth]{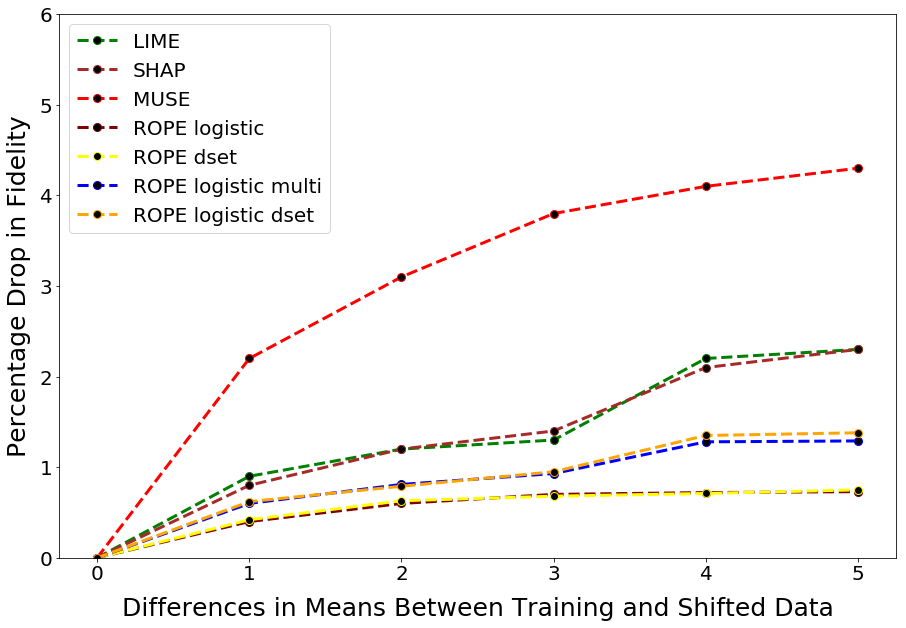}
\includegraphics[width=0.31\textwidth]{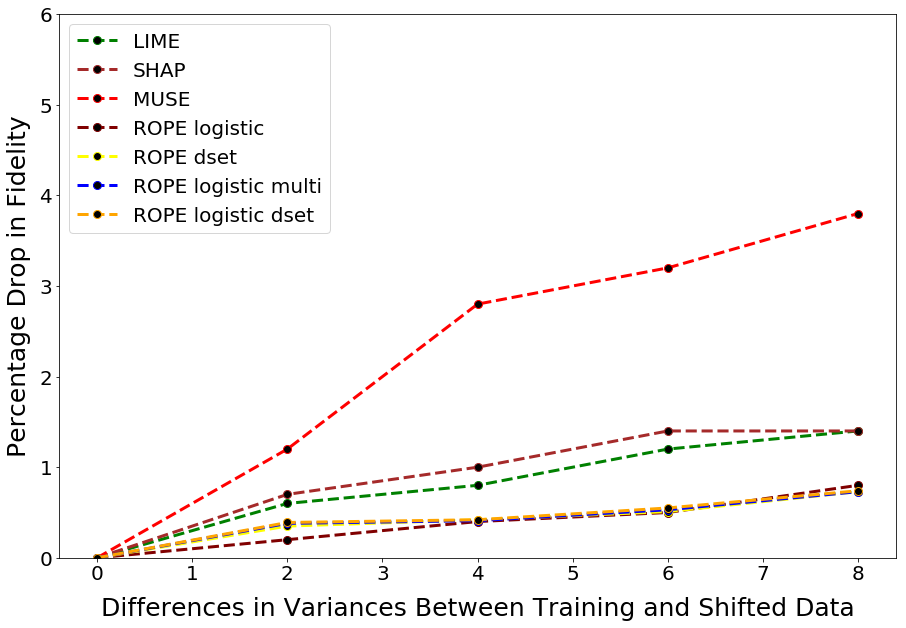}
\caption{Impact of changes in covariate correlations (left), means (middle), and variances (right) on percentage drop in fidelities. Lower values of percentage drop indicate higher robustness. Standard errors too small to be included.} 
\vspace{-0.2in}
\label{fig:analysis}
\end{figure*}

\textbf{Baselines.}
We compare our framework to the following state-of-the-art post hoc explanation techniques: (i) 
LIME~\cite{ribeiro16:kdd}, (ii)
SHAP~\cite{lundberg17:a-unified}, and (iii)
MUSE~\cite{lakkaraju19faithful}. LIME and SHAP are \emph{model-agnostic}, \emph{local explanation} techniques that explain an individual prediction of a black box by training a linear model on data near that prediction. LIME and SHAP can be adapted to produce global explanations of any given black box using a submodular pick procedure~\cite{ribeiro16:kdd}, which chooses a few representative points from the dataset and combines their corresponding local models to form a global explanation. In our evaluation, we use the global explanations of LIME and SHAP constructed using this technique. MUSE is a \emph{model-agnostic}, \emph{global explanation} technique; it provides global explanations in the form of two-level decision sets.

\textbf{Parameters.}
In case of LIME, SHAP, \rope logistic multi, and \rope dset multi, there is a parameter $K$ which corresponds to the number of local explanations that need to be generated; $K$ can also be thought as the number of subgroups in the data. We use Bayesian Information Criterion (BIC) to choose $K$. For a given dataset, we use the same $K$ for all these techniques to ensure they construct explanations of the same size. For MUSE, we set all the parameters using the procedure in~\citet{lakkaraju19faithful}; to ensure these explanations are similar in size to the others, we fix the number of outer rules to be $K$. Finally, when using \rope to construct rule-based explanations, there is a term $\lambda$ in our objective (Eq.~\ref{eqn:opti}); we fix $\lambda = 5$.

\textbf{Black boxes.}
We generate post hoc explanations of deep neural networks (DNNs), gradient boosted trees, random forests, and SVMs. Here, we present results for a 5-layer DNN; remaining results are included in the Appendix. Results presented below are representative of those for other model families.

\textbf{Metrics.}
We use fidelity to measure performance---i.e., the fraction of inputs $x$ in the given dataset for which $\hat{E}(x)=B^*(x)$~\cite{lakkaraju19faithful}. Fidelity is straightforward to compute for MUSE, \rope logistic, and \rope dset since they construct an explanation in the form of a single interpretable model. However, the explanations constructed by LIME, SHAP, \rope logistic multi, and \rope dset multi consist of a collection of local models. In these cases, we need to determine which local model to use for each input $x$. By construction, each local model $\hat{E}_i$ is associated with a representative input $x_i$, for $i\in\{1,...,K\}$. Thus, we compute the distance $\|x-x_i\|$ for each $i$, and return $\hat{E}_{i^*}(x)$ where $x_{i^*}$ is closest to $x$.

\subsection{Robustness to Real Distribution Shifts}
\label{sec:expreal}
We assess the robustness of explanations constructed using each approach on real-world datasets. In particular, we compute the fidelity of the explanations on both the training data and the shifted data, as well as the percentage change between the two. A large drop in fidelity from the training data to the shifted data indicates that the explanation is not robust. Ideally, explanations should have high fidelity on both the training data (indicating it is a good approximation of the black box model) and on the shifted data (indicating it is robust to distribution shift).

Results for all three real-world datasets are shown in Table~\ref{tab:distshift}. As can be seen, all the explanations constructed using our framework \rope have a much smaller drop in fidelity (0\% to 7\%) compared to those generated using the baselines. These results demonstrate that our approach significantly improves robustness. MUSE explanations have the largest percentage drop (21\% to 23\%), likely because MUSE relies entirely on the training data. In contrast, both LIME and SHAP employ input perturbations when constructing explanations~\cite{ribeiro16:kdd,lundberg2017unified}, resulting in somewhat increased robustness compared to MUSE. Nevertheless, LIME and SHAP still demonstrate a considerable drop (13\% to 19\%), so they are still not very robust. The reason is because these approaches do not optimize a minimax objective that encodes robustness such as ours.
Thus, these results validate our approach.

In addition, Table~\ref{tab:distshift} shows the actual fidelities on both training data and shifted data. As can be seen, the fidelities of
\rope logistic and \rope dset are lower than the other approaches; these results are expected since \rope logistic and \rope dset only use a single logistic regression and a single decision set model, respectively, to approximate the entire black box. On the other hand, \rope logistic multi and \rope dset multi achieve fidelities that are equal or better than the other baselines. These results demonstrate that \rope achieves robustness without sacrificing fidelity on the original training distribution. Thus, our approach \emph{strictly outperforms} the baseline approaches.

\begin{table*}
\small
\centering
\begin{tabular}{l c c c c c c}
    \toprule
        \bf Algorithms & \multicolumn{6}{c}{\bf Black Boxes} \\
        \midrule
       & {\bf LR} & {\bf Multiple LR} & \multicolumn{2}{c}{\bf DS} & \multicolumn{2}{c}{\bf Multiple DS}\\ 
    & Coefficient & Coefficient & Rule & Feature & Rule & Feature\\
    & Mismatch & Mismatch & Match & Match & Match & Match\\
    \midrule
    LIME & 4.37 & 5.01 & -- & -- & -- & --\\
    SHAP & 4.28 & 4.96 & -- & -- & -- & --\\
    MUSE & -- & -- & 4.39 & 11.81 & 4.42 & 9.23\\
    \midrule
    \rope logistic & {\bf 2.69} & 4.73 & -- & -- & -- & --\\
    \rope dset & -- & -- & {\bf 6.23} & 15.87& 4.78 & 11.23\\
    \midrule
    \rope logistic multi & 2.70 & {\bf 2.93} & -- & -- & -- & -- \\
    \rope dset multi & -- & -- & 6.25 & {\bf 16.18} & {\bf 7.09} & {\bf 16.78}\\
    \bottomrule
    \end{tabular}
\caption{Correctness of explanations on the bail dataset: Smaller coefficient mismatch and larger rule/feature match are better---i.e., the explanation more closely resembles the black boxes. \rope dset multi and \rope logistic multi uniformly outperform all the baselines.}
\label{tab:correctness}
\end{table*}

\subsection{Impact of Degree of Distribution Shift on Fidelity}

Next, we assess how different kinds of distribution shifts impact the fidelity of explanations constructed using our framework and the baselines using synthetic data. We study the effects of three different kinds of shifts: (i) changes in the correlations between different components of the covariates, (ii) changes in the means of the covariates, and (iii) changes in the variances of the covariates.

\textbf{Shifts in correlation.}
We first describe our study for shifted data of type (i) above. We generate a synthetic dataset with 5K samples. The covariate dimension is randomly chosen between 2 and 10. Each data point is sampled $x\sim\mathcal{N}(\mu,\Sigma)$, where $\mu_i=0$, $\Sigma_{ii}=1$ and $\Sigma_{ij}=\beta$, where $\beta$ is uniformly random in $[-1,1]$---i.e., the correlation between any two components of the covariates is $\beta$.
The label for each data point is chosen randomly. We train a 5 layer DNN $B^*$ on this dataset, and construct explanations for $B^*$.

To generate shifted data, we generate a new dataset with the same approach as above but using a different correlation $\beta'=\beta+\alpha$, where we vary $\alpha$.
Then, we compute the percentage drop in fidelity of the explanations from the training data to each of the shifted datasets. We show results averaged over 100 runs in Figure~\ref{fig:analysis} (left); the $x$-axis shows $|\alpha|$, and the $y$-axis shows the percentage drop. 
As can be seen, MUSE exhibits the highest drop in fidelity, followed closely by LIME and SHAP. In contrast, the \rope explanations are substantially more robust, incurring less than a 10\% drop in fidelity.  

\textbf{Mean shifts.}
For shifts of type (ii) above, we follow the same procedure, except we use $\beta=0$ for both the training and shifted datasets (i.e., uncorrelated covariates), and choose $\mu$ randomly in $[-5,5]$.
To generate shifted data, we use a different $\mu'=\mu+\alpha$.
Results averaged across 100 runs are shown in Figure~\ref{fig:analysis} (middle).
\rope is still the most robust, though LIME and SHAP are closer to \rope than to MUSE. Explanations generated by MUSE are not robust even to small changes in covariate means.

\textbf{Variance shifts.}
For shifts of type (iii) above, we follow the same procedure, except we use $\beta=0$, and choose $\Sigma_{ii}=\sigma$, where $\sigma$ is randomly chosen from $[1, 10]$. To generate shifted data, we use a different $\sigma'=\sigma+\alpha$.
Results averaged across 100 runs are shown in Figure~\ref{fig:analysis} (right). The results are similar to the case of mean shifts.

\subsection{Evaluating Correctness of Explanations}
Here, we evaluate the \emph{correctness} of the constructed explanations---i.e., how closely an explanation resembles the black box. To this end, we first train ``black box'' models $B^*\in\Finterp$ that are interpretable using the \emph{training data} from each of our real-world datasets.
Then, we construct an explanation $\hat{E}$ for $B^*$ using the \emph{shifted data}. If $\hat{E}$ resembles $B^*$ structurally, then the underlying explanation technique is generating explanations that are correct despite being constructed based on shifted data.

\textbf{Logistic regression black box.}
We first train a logistic regression (\textbf{LR}) ``black box" $B^*$, and then use LIME, SHAP, \rope logistic, and \rope logistic multi to construct explanations $\hat{E}$ for $B^*$. We define the \emph{coefficient mismatch} to measure the correctness. For \rope logistic, it is computed as $\|\hat{E}-B^*\|$---i.e., the $L_2$ distance between the weight vectors of $\hat{E}$ and $B^*$; smaller distances mean the explanation more closely resembles the black box. The remaining approaches construct multiple logistic regression models---one $\hat{E}_i$ for each representative input $x_i$, for $i\in\{1,...,K\}$.
To measure the coefficient mismatch, we assign a weight $w_i$ to each $x_i$ that equals the fraction of inputs $x$ that are assigned to $x_i$ (i.e., $x_i$ is the closest representative). Then, we measure coefficient mismatch as $\sum_{i=1}^Kw_i\cdot\|\hat{E}_i-B^*\|$.

We also consider the case where $B^*$ is a collection of multiple logistic regression (\textbf{Multiple LR})  models---one $B^*_i$ for each of the $K$ subgroups. We construct explanations using LIME, SHAP, \rope logistic, and \rope logistic multi, and measure the coefficient mismatch as $\sum_{i=1}^Kw_i\cdot\|\hat{E}_i-B^*_e\|$; In case of \rope logistic, $\hat{E}_i = \hat{E}$. 

Results for the bail dataset are shown in Table~\ref{tab:correctness}. When $B^*$ is a single logistic regression (LR), \rope logistic and \rope logistic multi explanations achieve the best performance and are about 38.2\% more structurally similar to $B^*$ than the baselines. When $B^*$ is multiple logistic regressions (Multiple LR), the coefficient mismatch of \rope logistic multi is at least 38.05\% lower than the baselines. We obtained similar results for the academic and health datasets. 

\textbf{Decision set black box.}
As before, we train a decision set (\textbf{DS}) ``black box'' $B^*$ on the real-world training data, and then construct an explanation $\hat{E}$ based on the shifted data using MUSE, \rope dset, and \rope dset multi. We consider two measures of correctness for \rope dset: (i) \emph{rule match}: the number $d_r(\hat{E},B^*)$ of rules present in both $\hat{E}$ and $B^*$, and (ii) \emph{feature match}: the number of features $d_f(\hat{E},B^*)$ present in both $\hat{E}$ and $B^*$. As before, for \rope dset multi and MUSE, we use the weighted measure $\sum_{i=1}^Kw_i\cdot d(\hat{E}_i,B^*)$, where $d=d_r$ and $d=d_f$ for rule match and feature match respectively. Higher rule and feature matches indicate that $\hat{E}$ better resembles $B^*$. We also consider the case where $B^*$ consists of multiple decision sets (\textbf{Multiple DS})---one $B^*_i$ for each of the $K$ subgroups.

On the bail dataset, \rope dset multi has 42.3\% (resp., 60.4\%) higher rule match than MUSE when $B^*$ corresponds to DS (resp., Multiple DS), and has at least 37\% higher feature match than the baselines.

\subsection{Evaluating Stability of Explanations}

Finally, we evaluate the \emph{stability} of the constructed explanations---i.e., how much do the explanations change if the input data is perturbed by a small amount. To this end, we first generate a synthetic dataset $D$ with 5000 samples as described in Section 4.3. Then, we generate the perturbed dataset $D'$ by adding a small amount 
of Gaussian noise to  data points---i.e., $x' = x + \epsilon$, where $\epsilon \sim \mathcal{N}(0, 0.2)$.\footnote{We experimented with other choices of variance in the range $(0.2, 1.0]$ and found similar results.} 
We then train LR, Multiple LR, DS, and Multiple DS  ``black boxes" $B^*$, and use LIME, SHAP, \rope logistic, \rope logistic multi, \rope dset, \rope dset multi to construct explanations $\hat E$ for the corresponding $B^*$. We use the original dataset $D$ both to train each black box $B^*$ as well as to construct its explanation $\hat E$. Then, for each black box $B^*$, we use the perturbed dataset $D'$ to construct an additional explanation $\hat E'$. Since $D'$ is obtained by making small changes to instances in $D$, $E$ and $E'$ should be structurally similar if the explanation technique used to construct them generates stable explanations.

We measure structural similarity of $E$ and $E'$---similar to the results in Table 2, we compute their coefficient mismatch in the case of LR and Multiple LR, and rule and feature match in case of DS and Multiple DS. We find that explanations $E$ and $E'$ constructed using \rope are 18.21\% to 21.08\% more structurally similar than those constructed using LIME, SHAP, or MUSE. Thus, our results demonstrate that \rope explanations are much more stable than those constructed using baselines.

\vspace{-0.1in}
\section{Conclusions \& Future Work}
In this paper, we proposed a novel framework based on adversarial training for constructing explanations that are robust to distribution shifts and are stable. Experimental results have demonstrated that our framework can be used to construct explanations that are far more robust to distribution shifts than those constructed using other state-of-the-art techniques. Our work paves way for several interesting future research directions. First, it would be interesting to extend our techniques to other classes of explanations such as saliency maps. Second, it would also be interesting to design adversarial attacks that can potentially exploit any vulnerabilities in our framework to generate unstable and incorrect explanations.  
\vspace{-0.1in}
\section*{Acknowledgements}
This work is supported in part by Google and NSF Award CCF-1910769. The U.S. Government is authorized to reproduce and distribute reprints for Governmental purposes notwithstanding any copyright notation thereon.

\bibliography{main}
\bibliographystyle{icml2020}
\clearpage
\appendix
\section{Additional Results}

\subsection{Robustness to Real Distribution Shifts}

We assess the robustness of explanations constructed using our approaches and the baselines on various real world datasets. The analysis that we present here is the same as that in Section 4.2, except for the underlying black boxes. In particular, we consider gradient boosted trees, random forests, and SVMs as black boxes. Corresponding results are presented in Tables~\ref{tab:gbt}, ~\ref{tab:rf}, and ~\ref{tab:svm} respectively. 

We observe similar results as that of Section 4.2 with other black boxes.
All the explanations constructed using our framework \rope have a much smaller drop in fidelity (0\% to 5\%) compared to those generated using the baselines. These results demonstrate that our approach significantly improves robustness. MUSE explanations have the largest percentage drop (13\% to 26\%). In contrast, both LIME and SHAP employ input perturbations when constructing explanations~\cite{ribeiro16:kdd,lundberg2017unified}, resulting in somewhat increased robustness compared to MUSE. Nevertheless, LIME and SHAP still demonstrate a considerable drop, so they are still not very robust. 
Thus, these results validate our approach.

Tables~\ref{tab:gbt}, ~\ref{tab:rf}, and ~\ref{tab:svm} also show the fidelities on both training data and shifted data. The fidelities of
\rope logistic and \rope dset are lower than the other approaches, which is expected since \rope logistic and \rope dset only use a single logistic regression and a single decision set, respectively, to approximate the entire black box. On the other hand, \rope logistic multi and \rope dset multi achieve fidelities that are equal or better than the other baselines. These results demonstrate that \rope achieves robustness without sacrificing fidelity on the original training distribution. Thus, our approach \emph{strictly outperforms} the baseline approaches.

\begin{table*}
\small
\centering
\begin{tabular}{l c c c c c c c c c}
    \toprule
        \bf Algorithms & \multicolumn{3}{c}{\bf Bail} & \multicolumn{3}{c}{\bf Academic} & \multicolumn{3}{c}{\bf Health} \\ \midrule
    & Train & Shift & \% Drop & Train & Shift & \% Drop & Train & Shift & \% Drop\\
    \midrule
    LIME & 0.73 & 0.61 & 16.31\% & 0.71 & 0.59 & 17.38\% & 0.78 & 0.67 & 14.31\% \\
    SHAP & 0.72 & 0.61 & 15.72\% & 0.69 &  0.58 & 16.37\% & 0.79 & 0.68 & 13.92\% \\
    MUSE & 0.69 & 0.57 & 18.02\% & 0.67 & 0.53 & 20.32\%  & 0.75 & 0.62 & 17.01\% \\
    \midrule
    \rope logistic & 0.59 & 0.57 & 3.02\% & 0.57 & 0.55 & 3.57\% & 0.68 & 0.66 & 2.32\% \\
    \rope dset & 0.63 & 0.61 & 2.98\% & 0.61 & 0.59 & 3.52\% & 0.74 & 0.73 & 1.92\%\\
    \midrule
    \rope logistic multi & 0.74 & 0.72 & 2.28\% & 0.71 & 0.69 & 2.45\% & 0.82 & 0.80  & 1.90\% \\
    \rope dset multi & {\bf 0.76} & {\bf 0.74} & {\bf 2.13\%} & {\bf 0.72} & {\bf 0.71} & {\bf 1.98\%} & {\bf 0.83} & {\bf 0.81} & {\bf 1.89\%} \\
    \bottomrule
    \end{tabular}
\caption{Gradient Boosted Trees (100 trees) as the black box. Fidelity values of all the explanations are reported on both training data and shifted data, along with percentage drop in fidelity from training data to shifted data. Smaller values of percentage drop correspond to more robust explanations.}
\label{tab:gbt}
\end{table*}

\begin{table*}
\small
\centering
\begin{tabular}{l c c c c c c c c c}
    \toprule
        \bf Algorithms & \multicolumn{3}{c}{\bf Bail} & \multicolumn{3}{c}{\bf Academic} & \multicolumn{3}{c}{\bf Health} \\ \midrule
    & Train & Shift & \% Drop & Train & Shift & \% Drop & Train & Shift & \% Drop\\
    \midrule
    LIME & 0.77 & 0.66 & 14.38\% & 0.69 &  0.61 & 11.83\% & 0.79 & 0.70 & 10.83\% \\
    SHAP & 0.74 & 0.61 & 16.98\% & 0.67 &  0.58 & 12.82\% & 0.77 & 0.69 & 11.02\% \\
    MUSE & 0.72 & 0.58 & 19.02\% & 0.65 & 0.55 & 15.01\% & 0.74 & 0.64 & 13.93\% \\
    \midrule
    \rope logistic & 0.63 & 0.62 & 2.32\% & 0.61 & 0.60 & 1.64\% & 0.69 & 0.68 & {\bf 1.59\%} \\
    \rope dset & 0.65 & 0.64 & 1.97\% & 0.63 & 0.62  &  {\bf 1.02\%} & 0.70 & 0.69 & 1.61\% \\
    \midrule
    \rope logistic multi & 0.78 & 0.76 & 2.38\% & 0.73 & 0.71 & 3.12\%  & 0.83 & 0.81 & 2.83\%\\
    \rope dset multi & {\bf 0.79} & {\bf 0.77} & {\bf 1.92\%} & {\bf 0.77} & {\bf 0.75} & 2.03\% & {\bf 0.86} & {\bf 0.84} &  1.77\% \\
    \bottomrule
    \end{tabular}
\caption{Random Forests (100 trees) as  the black box. Fidelity values of all the explanations are reported on both training data and shifted data, along with percentage drop in fidelity from training data to shifted data. Smaller values of percentage drop correspond to more robust explanations.}
\label{tab:rf}
\end{table*}

\begin{table*}
\small
\centering
\begin{tabular}{l c c c c c c c c c}
    \toprule
        \bf Algorithms & \multicolumn{3}{c}{\bf Bail} & \multicolumn{3}{c}{\bf Academic} & \multicolumn{3}{c}{\bf Health} \\ \midrule
    & Train & Shift & \% Drop & Train & Shift & \% Drop & Train & Shift & \% Drop\\
    \midrule
    LIME & 0.87 & 0.71 & 18.32\% & 0.89 &  0.74 & 17.27\% & 0.93 & 0.75 & 19.28\% \\
    SHAP & 0.87 & 0.73 & 16.32\% & 0.91 &  0.76 & 15.98\% & 0.93 & 0.79 & 15.56\%\\
    MUSE & 0.86 & 0.64 & 25.32\% & 0.87 &  0.67 & 23.41\% & 0.88 & 0.69 & 21.08\%\\
    \midrule
    \rope logistic & 0.81 & 0.79 & 2.39\%  & 0.84 & 0.83 &  {\bf 1.08\%} & 0.87 & 0.86 &  {\bf 0.98\%} \\
    \rope dset & 0.84 & 0.82 & 2.50\% & 0.86 & 0.84 & 2.32\% & 0.89 & 0.86 & 2.98\% \\
    \midrule
    \rope logistic multi & 0.89 & 0.87 & {\bf 1.98\%} & 0.92 & 0.89 & 3.32\% & 0.95 &  0.91 & 3.92\% \\
    \rope dset multi & {\bf 0.93} & {\bf 0.91} & 2.08\% & {\bf 0.93} & {\bf 0.90} & 3.32\% & {\bf 0.96} & {\bf 0.92} & 4.31\%  \\
    \bottomrule
    \end{tabular}
\caption{SVM as the black box. Fidelity values of all the explanations are reported on both training data and shifted data, along with percentage drop in fidelity from training data to shifted data. Smaller values of percentage drop correspond to more robust explanations.}
\label{tab:svm}
\end{table*}

\subsection{Impact of Degree of Distribution Shift on Fidelity}

\begin{figure*}
\centering
\includegraphics[width=0.31\textwidth]{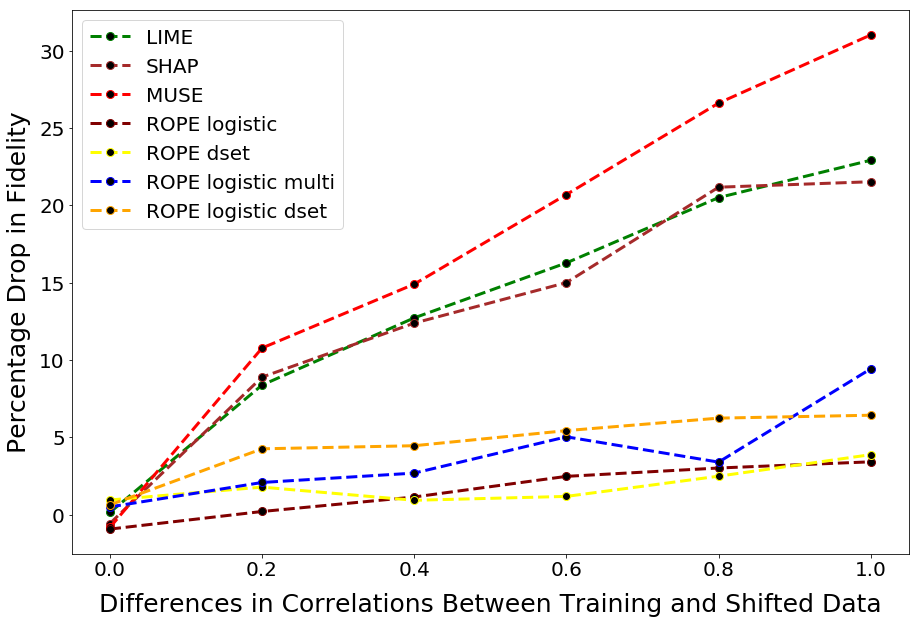}
\includegraphics[width=0.31\textwidth]{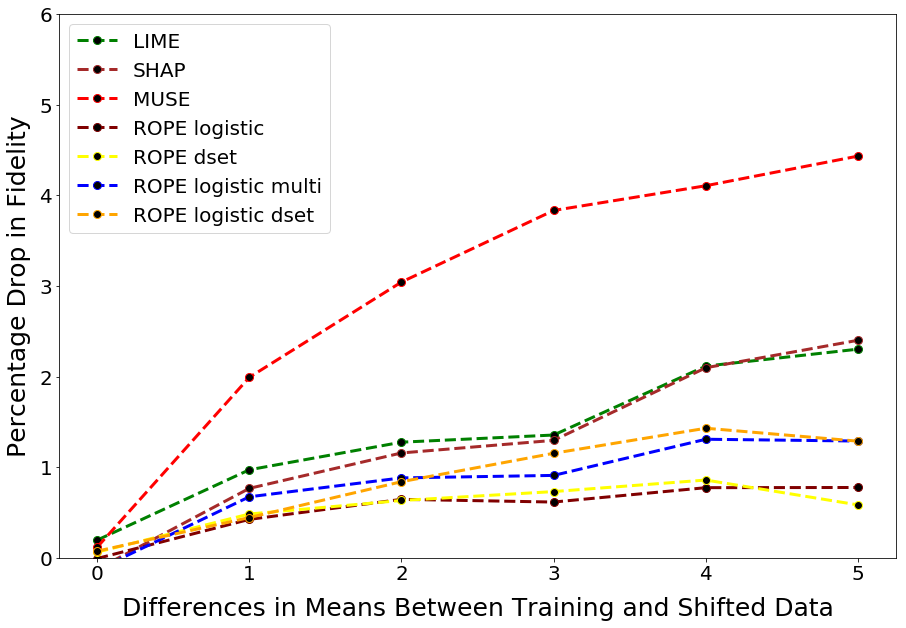}
\includegraphics[width=0.31\textwidth]{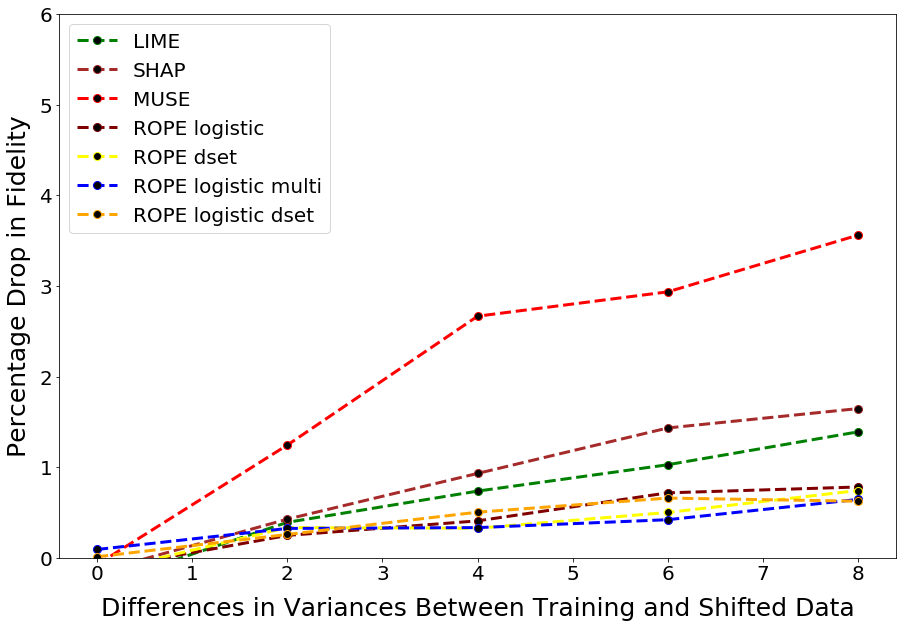}
\caption{Gradient Boosted Trees (100 trees) as the black box. Impact of changes in covariate correlations (left), means (middle), and variances (right) on percentage drop in fidelities. Lower values of percentage drop indicate higher robustness. Standard errors are too small to be included.} 
\label{fig:gbt}
\end{figure*}

\begin{figure*}
\centering
\includegraphics[width=0.31\textwidth]{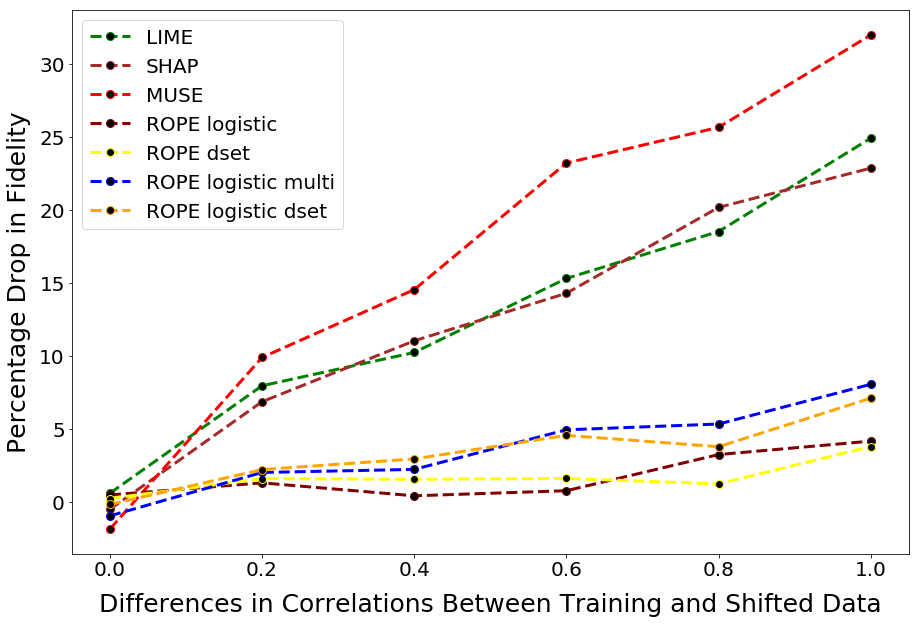}
\includegraphics[width=0.31\textwidth]{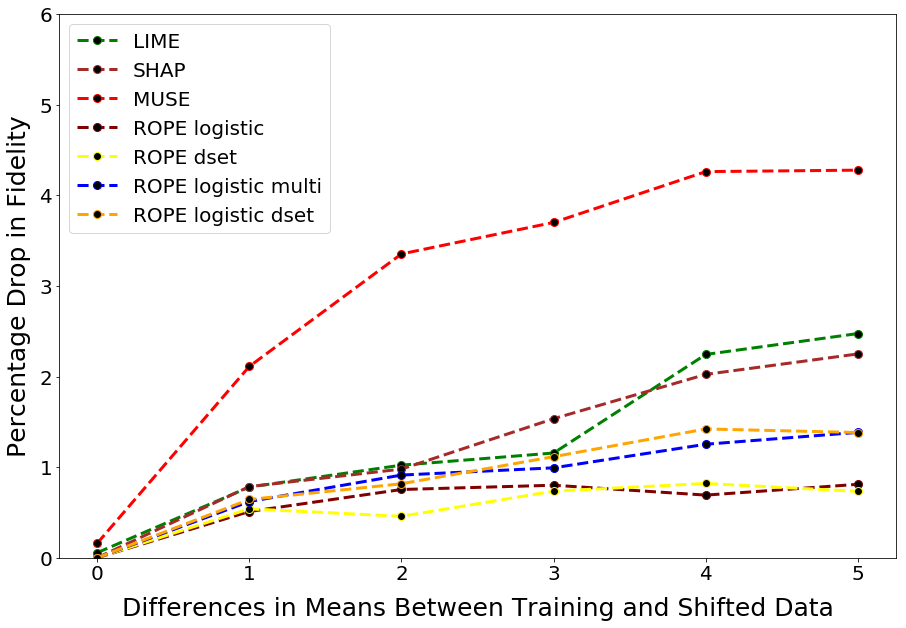}
\includegraphics[width=0.31\textwidth]{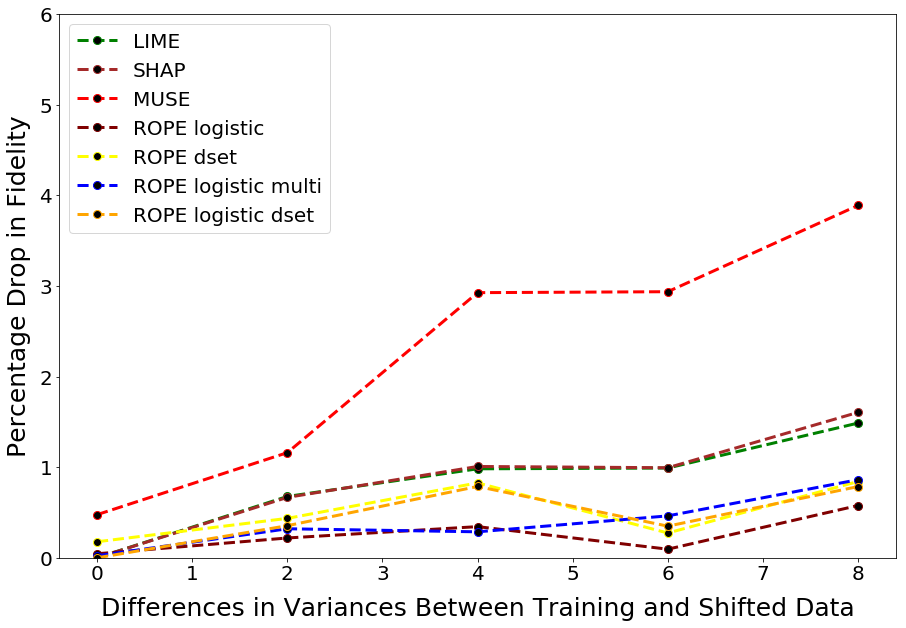}
\caption{Random Forests (100 trees) as the black box. Impact of changes in covariate correlations (left), means (middle), and variances (right) on percentage drop in fidelities. Lower values of percentage drop indicate higher robustness. Standard errors are too small to be included.} 
\label{fig:rf}
\end{figure*}

\begin{figure*}
\centering
\includegraphics[width=0.31\textwidth]{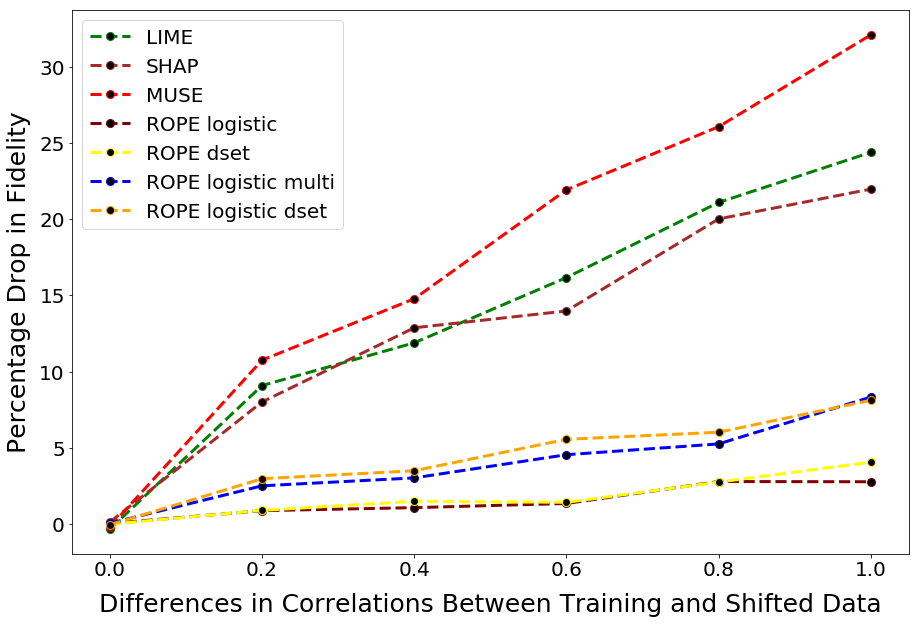}
\includegraphics[width=0.31\textwidth]{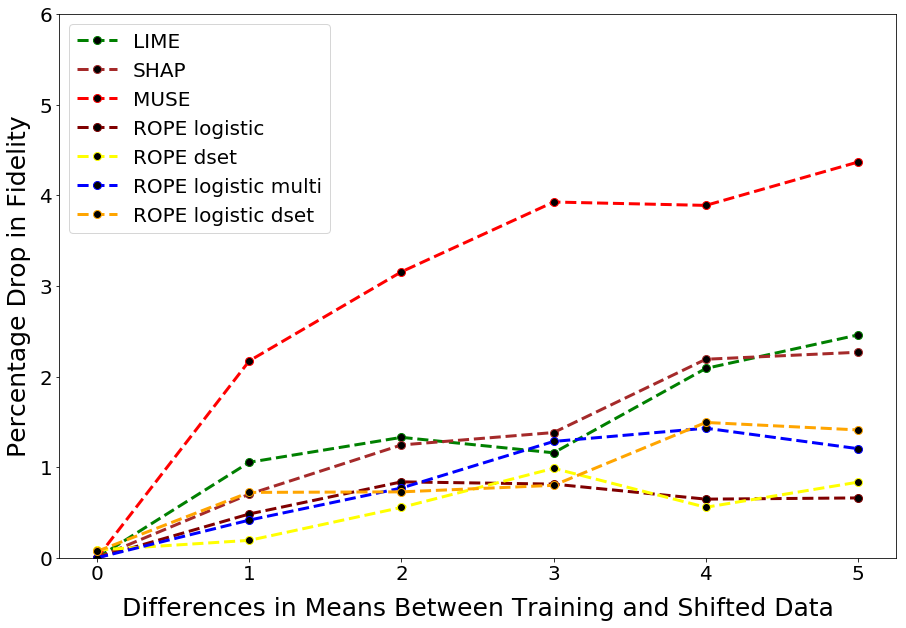}
\includegraphics[width=0.31\textwidth]{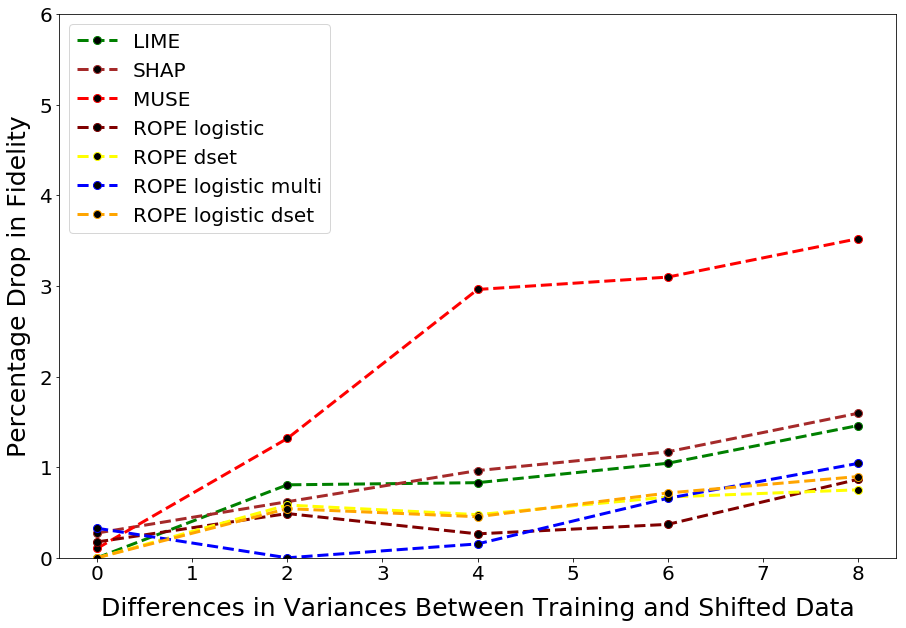}
\caption{SVM as the black box. Impact of changes in covariate correlations (left), means (middle), and variances (right) on percentage drop in fidelities. Lower values of percentage drop indicate higher robustness. Standard errors are too small to be included.} 
\label{fig:svm}
\end{figure*}

We replicate the analysis in Section 4.3, but with different black boxes. In particular, we consider gradient boosted trees, random forests, and SVMs as black boxes. Results are shown in Figures~\ref{fig:gbt}, ~\ref{fig:rf}, and ~\ref{fig:svm}, respectively. We observe similar patterns and trends as in Section 4.3.
\clearpage

\hide{
\section{Datasets \& Code}
The code for our approaches and MUSE baseline are included in the github link provided. In case of LIME and SHAP baselines, we used publicly available source codes~\cite{ribeiro16:kdd,lundberg17:a-unified} during experimentation. 

We also included two of the datasets (bail outcomes and academic performance) used in our experiments in the supplementary material. Our third dataset (electronic health records) has confidential patient information and, therefore, cannot be shared at this time. 

Note that we plan to make the first two datasets and code publicly available at the time of publication. }
\end{document}